\documentclass[11pt,a4paper]{article}
\usepackage[hyperref]{emnlp2020}
\usepackage{times}
\usepackage{latexsym}

\usepackage{microtype}

\usepackage{bm}
\usepackage{amsmath}
\usepackage{graphicx}
\usepackage{booktabs}
\usepackage{caption,subcaption}
\usepackage{multirow}
\usepackage{blindtext}
\usepackage{makecell}
\usepackage{enumitem}
\usepackage{amsfonts}
\usepackage{breqn}
\usepackage[ruled,vlined]{algorithm2e}
\usepackage{arydshln}

\makeatletter
\def\adl@drawiv#1#2#3{%
        \hskip.5\tabcolsep
        \xleaders#3{#2.5\@tempdimb #1{1}#2.5\@tempdimb}%
                #2\z@ plus1fil minus1fil\relax
        \hskip.5\tabcolsep}
\newcommand{\cdashlinelr}[1]{%
  \noalign{\vskip\aboverulesep
           \global\let\@dashdrawstore\adl@draw
           \global\let\adl@draw\adl@drawiv}
  \cdashline{#1}
  \noalign{\global\let\adl@draw\@dashdrawstore
           \vskip\belowrulesep}}
\makeatother

\aclfinalcopy 

\newcommand{\expnumber}[2]{{#1}\mathrm{e}{#2}}

\title{Meta-Learning with Sparse Experience Replay for Lifelong Language Learning}

\author{Nithin Holla \\
  ILLC, University of Amsterdam \\
  \texttt{nithin.holla7@gmail.com} \\\And
  Pushkar Mishra \\
  Facebook AI\\
  \texttt{pushkarmishra@fb.com} \\\AND
  Helen Yannakoudakis \\
  Dept. of Informatics, King's College London \\
  \texttt{helen.yannakoudakis@kcl.ac.uk} \\ \And
  Ekaterina Shutova \\
  ILLC, University of Amsterdam \\
  \texttt{e.shutova@uva.nl} \\}

\date{}

\begin{document}

\maketitle

\begin{abstract}
Lifelong learning requires models that can continuously learn from sequential streams of data without suffering catastrophic forgetting due to shifts in data distributions. Deep learning models have thrived in the non-sequential learning paradigm; however, when used to learn a sequence of tasks, they fail to retain past knowledge and learn incrementally. We propose a novel approach to lifelong learning of language tasks based on meta-learning with sparse experience replay that directly optimizes to prevent forgetting. We show that under the realistic setting of performing a single pass on a stream of tasks and without any task identifiers, our method obtains state-of-the-art results on lifelong text classification and relation extraction. We analyze the effectiveness of our approach and further demonstrate its low computational and space complexity. 
\end{abstract}

\section{Introduction} 

The ability to learn tasks continuously during a lifetime and with limited supervision is a hallmark of human intelligence. This is enabled by efficient transfer of knowledge from past experience. On the contrary, when current deep learning methods are subjected to learning new tasks in a sequential manner, they suffer from catastrophic forgetting \citep{mccloskey-catastrophic,ratcliff-connectionist,french-catastrophic}, where previous information is lost due to the shift in data distribution. Non-stationarity is inevitable in the real world where data is continuously evolving. Thus, we need to design more robust machine learning mechanisms to deal with catastrophic interference. 

Lifelong learning, also known as continual learning \citep{thrun-lifelong}, aims at developing models that can continuously learn from a stream of tasks in sequence without forgetting existing knowledge but rather building on the information acquired by previously learned tasks in order to learn new tasks \cite{chen2018lifelong}. One conceptualization of this is to accelerate learning by positive transfer between tasks while minimizing interference with respect to network updates \citep{riemer-MER}. Manually-designed techniques for continual learning include regularization \citep{kirkpatrick-ewc} or gradient alignment \citep{lopez-gem,chaudhry-agem} to mitigate catastrophic forgetting, and have been shown effective in computer vision and reinforcement learning tasks. Meta-learning \citep{schmidhuber_thesis,bengio_metalearning,thrun_metalearning} has been applied in continual learning with the objective of learning new tasks continually with a relatively small number of examples per task \citep{javed_oml, beaulieu-anml} (in image classification) or in a traditional continual learning setup by interleaving with several past examples from a memory component, i.e. experience replay \citep{riemer-MER,obamuyide-vlachos-2019-rel-extract} (in image classification, reinforcement learning and language processing). 

In natural language processing (NLP), continual learning still remains relatively unexplored \citep{li-compositional}. Despite the success of large pre-trained language models such as BERT \citep{devlin-bert}, they still require considerable amounts of in-domain examples for training on new tasks and are prone to catastrophic forgetting \cite{yogatama}. Existing continual learning approaches to NLP tasks include purely replay-based methods \citep{wang-embedding_alignment,han-continual_rel,deMasson-episodic_memory}, a meta-learning based method \citep{obamuyide-vlachos-2019-rel-extract,wang-efficient} as well as a generative replay-based method \citep{sun-lamol}. 

Currently, most of the approaches to continual learning employ flawed experimental setups that have blind spots disguising certain weak points. By assuming explicit task identifiers, distinct output heads per task, multiple training passes over the sequence of tasks, and the availability of large training times as well as computational and memory resources, they are essentially solving an easier problem compared to true continual learning \citep{farquhar_robust}. Specifically, while a high rate of experience replay \citep{lin-experience_replay} usually mitigates catastrophic forgetting, it comes closer to a multi-task learning \citep{caruana-multitask} than a lifelong learning setup and is computationally expensive when learning on a data stream in real-life applications. Continual learning methods in NLP suffer from these limitations too. An appropriate evaluation of continual learning is thus one where no task identifiers are available, without multiple epochs of training, with a shared output head as well as constraints on time, compute and memory.

In this paper, we propose a novel and efficient approach to lifelong learning on language processing tasks that overcomes the aforementioned shortcomings. We consider the realistic lifelong learning setup where only one pass over the training set is possible with constraints on the rate of experience replay, and no task identifiers are available. Our approach is based on meta-learning and experience replay that is sparse in time and size. We are the first to investigate meta-learning with sparse experience replay in the context of large-scale pre-trained language models, in contrast with previous works that take liberty in terms of how often experience replay is performed. Additionally, we conduct a systematic study of approaches that rely on pre-trained models and that conform to the realistic lifelong learning setup.

We extend two algorithms, namely \textit{online meta-learning} (OML) \citep{javed_oml} and \textit{a neuromodulatory meta-learning algorithm} (ANML) \citep{beaulieu-anml} to the domain of NLP and augment them with an episodic memory module for experience replay, calling them OML-ER and ANML-ER respectively. While their original objective is to continually learn a new sequence of tasks during testing time, we enhance them for the conventional continual learning setup where evaluation is on previously seen tasks, thus directly addressing the problem of catastrophic forgetting. Furthermore, by realizing experience replay as a query set, we directly optimize to prevent forgetting. 

We show that combining a pre-trained language model such as BERT along with meta-learning and sparse replay produces state-of-the-art performance on lifelong text classification and relation extraction benchmarks when compared against current methods under the same realistic setting. 
Through further experiments, we demonstrate that BERT combined with OML-ER results in an efficient form of lifelong learning, where most of the weight updates are performed on a single linear layer on top of BERT, while using a limited amount of memory during training and without any network adaptation during test-time. Therefore, our approach is considerably more efficient than previous work in terms of computational complexity as well as memory usage, enabling learning on a task stream without substantial overheads. We hope that our paper informs the NLP community about the right experimental design of continual learning and how meta-learning methods enable efficient lifelong learning with limited replay and memory capacity. To facilitate further research in the field, we make our code publicly available\footnote{\url{https://github.com/Nithin-Holla/MetaLifelongLanguage}}.

\section{Background and Related Work}

\subsection{Meta-learning}

In meta-learning, a model is trained on several related tasks such that it can transfer knowledge and adapt to new tasks using only a few examples. The training set is referred to as \textit{meta-training set} and the test set is referred to as \textit{meta-test set}. They consist of \textit{episodes} where each episode corresponds to a task, comprising a few training examples for adaptation called the \textit{support set} and a separate set of examples for evaluation called the \textit{query set}. The goal of meta-learning is to learn to adapt quickly from the support set such that the model can perform well on the query set.

Optimization-based methods for meta-learning explicitly include generalizability in their objective function and optimize for the same. Model-agnostic meta-learning (MAML) algorithm \citep{finn} is an optimization-based method that seeks to train a model's initial parameters such that it can perform well on a new task after only a few gradient steps. During meta-training, it involves a two-level optimization process where task adaptation is performed using the support set in an \textit{inner-loop} and meta-updates are performed using the query set in an \textit{outer-loop}. Specifically, parameters $\bm{\theta}$ of the model $f_{\bm{\theta}}$ are updated to $\bm{\theta}_i'$ for task $\mathcal{T}_i$ in the inner-loop by $m$ steps of gradient-based update $U$ on the support set as:
\begin{equation} \label{eqn:inner_loop}
    \bm{\theta}_i' = U(\mathcal{L}_{\mathcal{T}_i}^s, \bm{\theta}, \alpha, m)
\end{equation}
where $\mathcal{L}_{\mathcal{T}_i}^s$ is the loss on the support set and $\alpha$ is the inner-loop learning rate. The outer-loop objective is to have $f_{\bm{\theta}_i'}$ generalize well across tasks from a distribution $p(\mathcal{T})$:
\begin{equation}
    J(\bm{\theta}) = \sum_{\mathcal{T}_i \sim p(\mathcal{T})} \mathcal{L}_{\mathcal{T}_i}^q (f_{U(\mathcal{L}_{\mathcal{T}_i}^s, \bm{\theta}, \alpha, m)})
\end{equation}
where $\mathcal{L}_{\mathcal{T}_i}^q$ is the loss computed on the query set. The outer-loop optimization does the update with the outer-loop learning rate $\beta$ as:
\begin{equation}
    \bm{\theta} \leftarrow \bm{\theta} - \beta \nabla_{\bm{\theta}} \sum_{\mathcal{T}_i \sim p(\mathcal{T})} \mathcal{L}_{\mathcal{T}_i}^q (f_{\bm{\theta}_i'})
\end{equation}

This involves computing second-order gradients, i.e., the backward pass works through the update step in Equation \ref{eqn:inner_loop}, which is a computationally expensive process. \citet{finn} propose a first-order approximation, called FOMAML, which computes the gradients with respect to $\bm{\theta}_i'$ rather than $\bm{\theta}$. The outer-loop optimization step thus reduces to:
\begin{align}
    \bm{\theta} \leftarrow \bm{\theta} - \beta \sum_{\mathcal{T}_i \sim p(\mathcal{T})} \nabla_{\bm{\theta}_i'} \mathcal{L}_{\mathcal{T}_i}^q (f_{\bm{\theta}_i'})
\end{align}

During meta-testing, new tasks are learned from the support sets and the performance is evaluated on the corresponding query sets.

Optimization-based meta-learning methods \citep{finn,Nichol,triantafillou_metadataset} have been shown to work well for few-shot learning problems in NLP -- specifically machine translation \citep{gu-etal-2018-meta}, relation classification \citep{obamuyide-vlachos-2019-rel-classification}, sentence-level semantic tasks \citep{dou-etal-meta,bansal_meta}, text classification \citep{jiang_text}, and word sense disambiguation \citep{holla}. 

\subsection{Continual learning} 

Current approaches to prevent catastrophic forgetting can be grouped into one of several categories: (1) constrained optimization-based approaches with or without regularization \citep{kirkpatrick-ewc,zenke-continual,Chaudhry-reimannian,aljundi-MAS,schwarz-distillation} that prevent large updates on weights that are important to previously seen tasks; (2) memory-based approaches \citep{rebuffi-icarl,sprechmann-mbpa,wang-embedding_alignment,deMasson-episodic_memory} that replay examples stored in the memory; (3) generative replay-based approaches \citep{shin_generative, kemker-fearnet,sun-lamol} that employ a generative model instead of a memory module; (4) architecture-based approaches \citep{rusu-progressive,chen-net2net,fernando-pathnet} that either use different subsets of the network for different tasks or dynamically expand the networks; and (5) hybrid approaches that formulate optimization constraints based on examples in memory \citep{lopez-gem, chaudhry-agem}. More recently, \citet{riemer-MER} proposed an approach based on a first-order optimization-based meta-learning algorithm, Reptile \citep{Nichol}, augmented with experience replay. However, it involved interleaving every training example with several examples from memory, leading to a high replay rate. 

Elastic Weight Consolidation (EWC) \citep{kirkpatrick-ewc}, Gradient Episodic Memory (GEM) \citep{lopez-gem} and Averaged-GEM (A-GEM) \citep{chaudhry-agem} are three popular continual learning methods. EWC introduces a regularization term involving the Fisher information matrix that indicates the importance of each of the parameters to previous tasks. GEM solves a constrained optimization problem as a quadratic program involving gradients from all examples from previous tasks. A-GEM is a more efficient version of GEM since it solves a simpler constrained optimization problem based on gradients from randomly drawn samples from previous tasks in the memory.

\subsection{Continual learning in NLP}
\citet{wang-embedding_alignment} propose an alignment model named EA-EMR that limits the distortion in the embedding space in an LSTM-based \citep{hochreiter_lstm} architecture for lifelong relation extraction. For the same task, \citet{obamuyide-vlachos-2019-rel-extract} show that utilizing Reptile \citep{Nichol} with memory can improve performance and call their method MLLRE. \citet{han-continual_rel} further improve relation extraction with their model, EMAR, through episodic memory activation and reconsolidation. \citet{deMasson-episodic_memory} propose a model with episodic memory called MbPA++ which incorporates sparse experience replay during training and local adaptation on $K$-nearest neighbors from the memory for every example during inference. Through their experiments on sequential learning on multiple datasets of text classification and question answering with BERT, they show that their model can effectively reduce catastrophic forgetting. Meta-MbPA \citep{wang-efficient} incorporates local adaptation during meta-training and performs fewer local adaptation steps during testing. \citet{sun-lamol} present a model based on GPT-2 \citep{radford-gpt2}, called LAMOL, that simultaneously learns to solve new tasks and to generate pseudo-samples from previous tasks for replay. They perform sequential learning on five tasks from decaNLP \citep{mccann-decaNLP} as well as multiple datasets for text classification. 

All these methods are not yet well-suited for application in real-life scenarios -- MbPA++ has slow inference, Meta-MbPA has a higher rate of sampling examples from memory, and other methods require task identifiers and multiple epochs of training. Our approach, on the other hand, alleviates all these problems. 

\section{Overview}

\subsection{Task formulation} 

A typical continual learning setup consists of a stream of $K$ tasks $\mathcal{T}_1$, $\mathcal{T}_2$, ..., $\mathcal{T}_K$. For supervised learning tasks, every task $\mathcal{T}_i$ consists of a set of data points $x_j$ with labels $y_j$, i.e., $\{(x_j, y_j)\}_{j=1}^{N_i}$ that are locally i.i.d., where $N_i$ is the size of task $\mathcal{T}_i$. We consider the setting where the goal is to learn a function $f_{\bm{\theta}}$ with parameters $\bm{\theta}$ by only making one pass over the stream of tasks and with no identifiers of tasks $\mathcal{T}_i$ available. In multi-task learning, on the other hand, it is possible to draw samples i.i.d from all tasks along with training for multiple epochs. Therefore, multi-task learning is an upper bound to continual learning in terms of performance. 

We propose an approach to continual learning with meta-learning and experience replay where the updates are similar to first-order MAML. We maintain an episodic memory (or simply called memory) $\mathcal{M}$ which stores previously seen examples. Episodes for meta-training are constructed from the stream of examples as well as randomly sampled examples from $\mathcal{M}$. We perform experience replay sparsely, i.e., a small number of examples are drawn from $\mathcal{M}$ and only after seeing many examples from the stream (i.e. at long intervals), therefore being computationally inexpensive.

\subsection{Motivation} \label{sec:motivation}
\citet{riemer-MER} note that, given two sets of gradients for shared parameters $\bm{\theta}$, interference occurs when the dot product of gradients is negative, and transfer occurs when their dot product is positive. Additionally, they show that Reptile \citep{Nichol} implicitly maximizes the dot product between gradients within an episode and, hence, when coupled with experience replay, it could facilitate continual learning. 

Consider a first-order MAML setup that performs one step of SGD on each of the $m$ batches in the support set during the inner-loop of an episode. Starting with parameters $\bm{\theta}_0 = \bm{\theta}$, it results in a sequence of parameters $\bm{\theta}_1$, ..., $\bm{\theta}_m$ using the losses $\mathcal{L}^1$, ..., $\mathcal{L}^m$. The meta-gradient computed on the query set of the episode is:
\begin{equation}
    g_{\text{FOMAML}} = \frac{\partial \mathcal{L}^q(\bm{\theta}_m)}{\partial \bm{\theta}_m}
\end{equation}
Using Taylor series approximation as in \citet{Nichol}, the expected gradient under mini-batch sampling could be expressed as:

\small
\begin{multline}
    \mathbb{E} \left[ g_{\text{FOMAML}} \right] = \mathbb{E} \left[ \frac{\partial \mathcal{L}^q(\bm{\theta}_m)}{\partial \bm{\theta}} - \frac{\alpha}{2} \frac{\partial}{\partial \bm{\theta}} \left( \sum_{j=1}^{m} \frac{\partial \mathcal{L}^j(\bm{\theta}_{j-1})}{\partial \bm{\theta}} \cdot \right. \right. \\
    \left. \left. \frac{\partial \mathcal{L}^q(\bm{\theta}_m)}{\partial \bm{\theta}} \right) \right] + O(\alpha^2)
\end{multline}
\normalsize
where $\alpha$ is the inner-loop learning rate. We provide a more detailed derivation in Appendix \ref{sec:grad_deriv}. Outer-loop gradient descent with this gradient approximately solves the following optimization problem:

\small
\begin{equation}
    \min_{\theta} \mathbb{E} \left[ \mathcal{L}^q(\bm{\theta}_m) - \frac{\alpha}{2} \left( \sum_{j=1}^{m} \frac{\partial \mathcal{L}^j(\bm{\theta}_{j-1})}{\partial \bm{\theta}} \cdot \frac{\partial \mathcal{L}^q(\bm{\theta}_m)}{\partial \bm{\theta}} \right) \right]
\end{equation}
\normalsize

This objective seeks to minimize the loss on the query set along with maximizing the dot product between the support and query set gradients. Thus, integrating previously seen examples into the query set in a first-order MAML framework could also potentially improve continual learning by minimizing interference and maximizing transfer.
 
\subsection{Episode generation and experience replay}

\begin{figure*}
    \centering
    \includegraphics[scale=0.4]{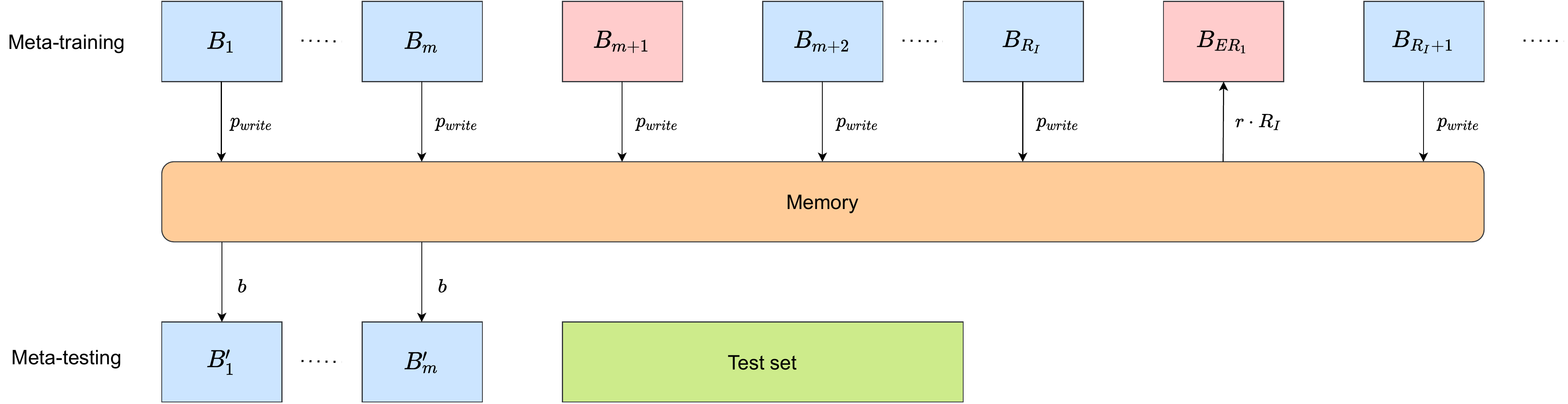}
    \caption{Illustration of meta-learning for lifelong learning. The $m$ mini-batches that form the support set used for inner-loop optimization are shown in blue. Red boxes indicate query sets used in outer-loop optimization. After every $R_I$ examples, the query set is obtained by sampling $r \cdot R_I$ examples from the memory, whereas other query sets are derived from the data stream. During meta-testing, $m$ mini-batches are sampled from the memory for fine-tuning, followed by evaluation on the test set.}
    \label{fig:stream}
\end{figure*}

We assume that data points arrive in mini-batches of a given size $b$ and every data point has a probability $p_{write}$ of being written into an episodic memory module $\mathcal{M}$. We construct episodes on-the-fly from the stream of mini-batches. Given a buffer size $m$, we construct episode $i$ by taking $m$ mini-batches as the support set $\mathcal{S}_i$ and the next batch as the query set $\mathcal{Q}_i$.

We explicitly define our experience replay mechanism as consisting of two fixed hyperparameters -- replay interval $R_I$, which indicates the number of data points seen between two successive draws from memory, and replay rate $r \in [0, 1]$ which indicates the proportion of examples to draw from memory relative to $R_I$. Thus, after every $R_I$ examples from the stream, $\lfloor r \cdot R_I \rfloor$ examples are drawn from the memory. 

We use these sampled examples from memory as the query set. To perform experience replay in an episodic fashion, we compute the replay frequency $R_F$ as follows (see Appendix \ref{sec:RF} for further details): 
\begin{equation} \label{eqn:replay_frequency}
    R_F = \left \lceil \frac{R_I / b + 1}{m + 1} \right \rceil
\end{equation}
Hence, every $R_F$ episodes, we draw a random batch of size $\lfloor r \cdot R_I \rfloor$ from $\mathcal{M}$ as the query set. For other episodes, the query set is obtained from the data stream. The support set for replay episodes is still constructed from the stream.  A high $r$ and/or a low $R_I$ ensures that information is not forgotten, but in order to be computationally efficient and adhere to continual learning, it is necessary that $r$ is low and $R_I$ is high, so that the replay is sparsely performed, both in terms of size and time.

During meta-testing, we randomly draw $m$ batches from the memory as the support set and take the entire test set of the respective task as the query set for evaluation. This is done primarily in order to match the testing and training conditions \citep{Vinyals}. Figure \ref{fig:stream} provides an illustration of the structure of episodes and experience replay.

\section{Methods}

\subsection{OML-ER}

\begin{figure}[t]
    \centering
    \includegraphics[width=0.9\linewidth]{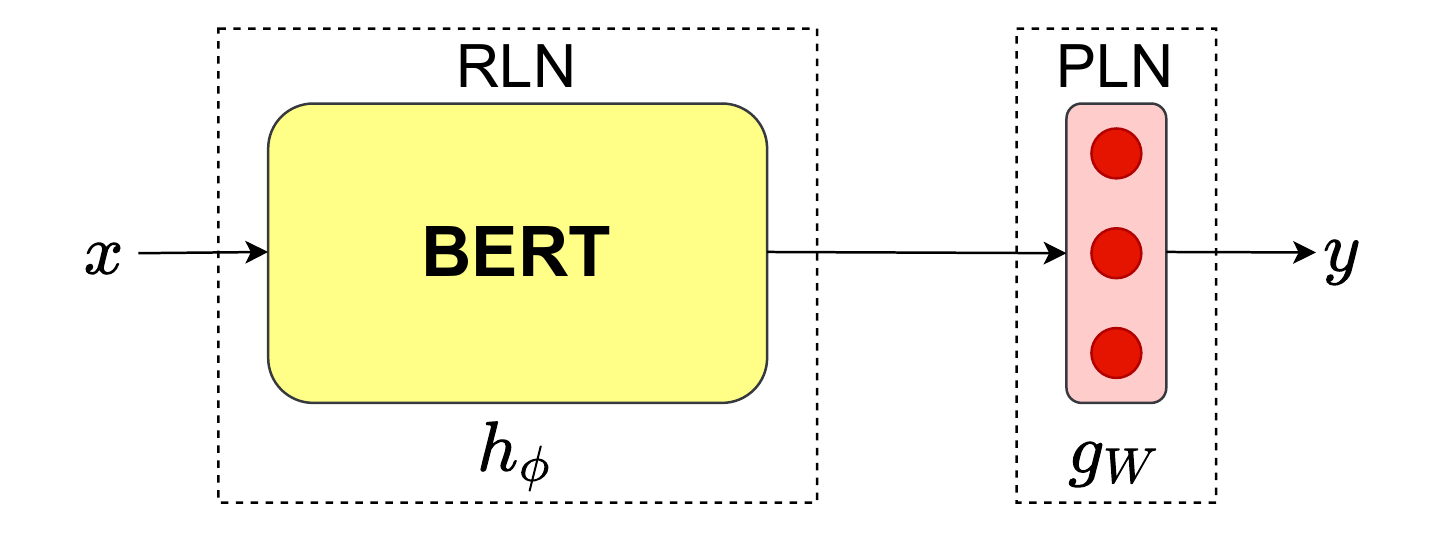}
    \caption{Architecture of OML.}
    \label{fig:arch_oml}
\end{figure}

The original OML algorithm \citep{javed_oml} was designed to solve new continual learning problems during meta-testing. Here, we extend it to our setup by augmenting it with an episodic memory module to perform experience replay (ER), and refer to it as OML-ER.

The model $f_{\bm{\theta}}$ is composed of two functions -- a representation learning network (RLN) $h_{\bm{\phi}}$ with parameters $\bm{\phi}$ and a prediction learning network (PLN) $g_{\bm{W}}$ with parameters $\bm{W}$ such that $\bm{\theta} = \bm{\phi} \cup \bm{W}$ and $f_{\bm{\theta}}(x) = g_{\bm{W}}(h_{\bm{\phi}}(x))$ for an input $x$. In each episode, the RLN is frozen while the PLN is fine-tuned during the inner-loop optimization. In the outer-loop, both the RLN and the PLN are meta-learned.

During the inner-loop optimization in episode $i$, the PLN is fine-tuned  on the support set mini-batches $\mathcal{S}_i$ with SGD to give:
\begin{equation} \label{eqn:inner_loop_oml}
    \bm{W}_i' = \text{SGD} \left( \mathcal{L}_i, \bm{\phi}, \bm{W}, \mathcal{S}_i, \alpha \right)
\end{equation}
where $\mathcal{L}_i$ is the loss function. Using the query set, the objective we optimize for is: 
\begin{equation} \label{eqn:regular_obj}
    J (\bm{\theta}) = \mathcal{L}_i \left( \bm{\phi}, \bm{W}_i', \mathcal{Q}_i \right)
\end{equation}
During a regular episode, the objective encourages generalization to unseen data  whereas during a replay episode, it promotes retention of knowledge from previously seen data. 

For the outer-loop optimization, we use the Adam optimizer \citep{kingma-adam} with a learning rate $\beta$ to update all parameters -- both the RLN and PLN:
\begin{equation} \label{eqn:outer_loop}
    \bm{\theta} \leftarrow \text{Adam} (J(\bm{\theta}), \beta)
\end{equation}
The above optimization would involve second-order gradients. Instead, we use the first-order variant where the gradients are taken with respect to $\bm{\theta}_i' = \bm{\phi} \cup \bm{W}_i'$.

We use $\text{BERT}_{\text{BASE}}$ \citep{devlin-bert} as the RLN (fully fine-tuned; output from the [CLS] token) and a single linear layer mapping to the classes as the PLN. The architecture of the model is shown in Figure \ref{fig:arch_oml}. 
 
\subsection{ANML-ER}

\begin{figure}[t]
    \centering
    \includegraphics[width=0.9\linewidth]{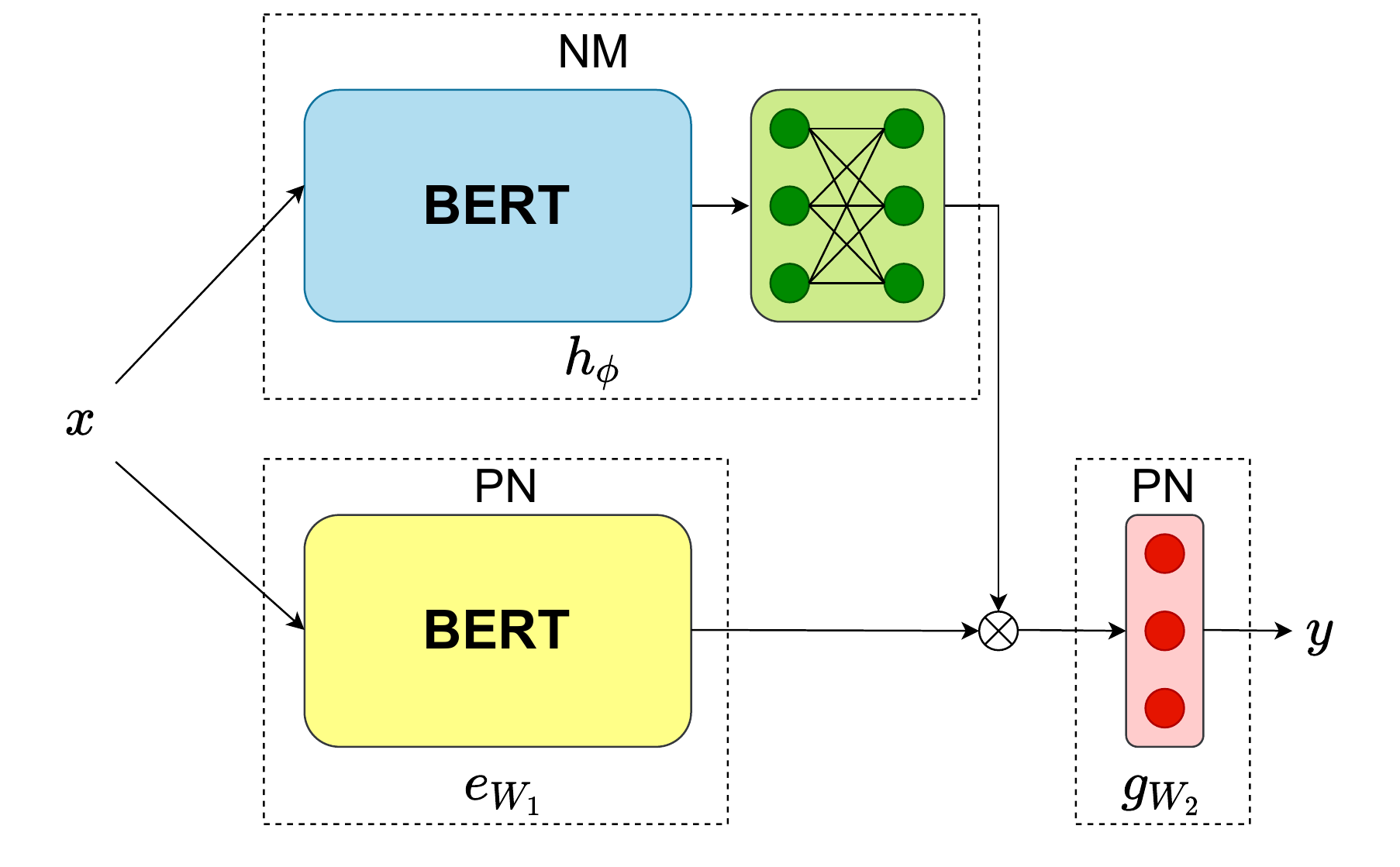}
    \caption{Architecture of ANML.}
    \label{fig:arch_anml}
\end{figure}

\citet{beaulieu-anml} proposed ANML that outperformed OML in solving new continual learning problems in image classification. Inspired by neuromodulatory processes in the brain, they design a context-dependent gating mechanism to achieve selective plasticity, i.e., limited and/or selective modification of parameters with new data. We refer to our extension of this method as ANML-ER.

The model $f_{\bm{\theta}}$ is composed of two networks -- a regular prediction network (PN) and a neuromodulatory network (NM) that selectively gates the internal activations of the prediction network via element-wise multiplication. Formally, the NM is a function $h_{\bm{\phi}}$ with parameters $\bm{\phi}$, and the PN is a composite function $g_{\bm{W}_2} \circ e_{\bm{W}_1}$ with parameters $\bm{W} = \bm{W}_1 \cup \bm{W}_2$. The output is obtained as:
\begin{equation}
    f_{\bm{\theta}} (x) = g_{\bm{W}_2} \left( e_{\bm{W}_1}(x) \cdot h_{\bm{\phi}}(x) \right)
\end{equation}

In the inner-loop, the NM is fixed while the PN is fine-tuned on the support set. Due to the choice of our notation, Equation \ref{eqn:inner_loop_oml} is the form of the inner-loop here too. In the outer-loop, both the NM and the PN are updated as in Equation \ref{eqn:outer_loop} with first-order gradients. 

Our PN is the $\text{BERT}_{\text{BASE}}$ encoder followed by a linear layer mapping to the classes as in OML-ER. For the NM, we use $\text{BERT}_{\text{BASE}}$ followed by two linear layers ($768$ units) with ReLU non-linearity between them and a final sigmoid non-linearity to limit the gating signal to $[0, 1]$. We keep the NM BERT frozen throughout to reduce the total number of parameters. Our preliminary experiments indicated that fine-tuning the NM BERT in addition produces negligible improvements. Figure \ref{fig:arch_anml} presents the model architecture. 

Algorithms \ref{algo:meta-train} and \ref{algo:meta-test} outline the meta-training and meta-testing procedure respectively that is common to both OML-ER and ANML-ER. 

\begin{algorithm}[ht]
\SetAlgoLined
\SetArgSty{textnormal}
\SetKwInput{Input}{Input}
\SetKwInput{Output}{Output}
\Input{Initial model parameters $\bm{\theta} = \bm{\phi} \cup \bm{W}$, replay interval $R_I$, replay frequency $R_F$, replay rate $r$, support set buffer size $m$, memory $\mathcal{M}$, write probability $p_{write}$, inner-loop learning rate $\alpha$, outer-loop learning rate $\beta$}
\Output{Trained model parameters $\bm{\theta}$, updated memory $\mathcal{M}$}
 \For{$i = 1, 2, ...$}{
    $\mathcal{S}_i \gets m$ batches from the stream\\
    \If{$i = R_F$}{
    $\mathcal{Q}_i \gets$ sample$(\mathcal{M}, \lfloor r \cdot R_I \rfloor)$\\
    }
    \Else{
    $\mathcal{Q}_i \gets$ next batch from the stream\\
    write$(\mathcal{M}, \mathcal{Q}_i, p_{write})$\\
    }
    write$(\mathcal{M}, \mathcal{S}_i, p_{write})$\\
    $\bm{W}_i' = \text{SGD}(\mathcal{L}_i, \bm{\phi}, \bm{W}, \mathcal{S}_i, \alpha)$\\
    $J (\bm{\theta}) = \mathcal{L}_i \left( \bm{\phi}, \bm{W}_i', \mathcal{Q}_i \right)$\\
    $\bm{\theta} \leftarrow \text{Adam} (J(\bm{\theta}), \beta)$
 }
\caption{Meta-training}
\label{algo:meta-train}
\end{algorithm}

\begin{algorithm}[ht]
\SetAlgoLined
\SetArgSty{textnormal}
\SetKwInput{Input}{Input}
\SetKwInput{Output}{Output}
\Input{Trained model parameters $\bm{\theta} = \bm{\phi} \cup \bm{W}$, support set buffer size $m$, memory $\mathcal{M}$, batch size $b$, inner-loop learning rate $\alpha$, test set $T$}
\Output{Predictions on the test set}
$\mathcal{S} \gets$ sample$(\mathcal{M}, m \cdot b)$\\
$\mathcal{Q} \gets T$ \\
$\bm{W}' = \text{SGD}(\mathcal{L}, \bm{\phi}, \bm{W}, \mathcal{S}, \alpha)$\\
predict$\left( \mathcal{Q}, \bm{\phi}, \bm{W}' \right)$\\
\caption{Meta-testing}
\label{algo:meta-test}
\end{algorithm}

\subsection{Baselines}

We consider four BERT-based baselines to evaluate the effectiveness of our approach.

\paragraph{SEQ} We train our model ``traditionally" on all tasks in a sequential manner i.e., one after the other, without replay. 

\paragraph{REPLAY} It is an extension of SEQ that incorporates sparse experience replay. After seeing $R_I$ examples from the stream, i.e., a replay frequency $R_F = \lceil R_I / b \rceil$,  $\lfloor r \cdot R_I \rfloor$ examples are randomly drawn from the memory and one gradient update is performed on them.

\paragraph{A-GEM} It requires replay at every training step and task identifiers by default \citep{chaudhry-agem}, but we adapt it to our setting by randomly sampling data points from the memory in sparse intervals.

\paragraph{MTL} We train our model in a ``traditional" multi-task setup for multiple epochs on mini-batches that are sampled i.i.d from the pool of all tasks. Thus, it serves as an upper bound for the performance of continual learning methods.

\section{Experimental setup}

\subsection{Datasets}

\paragraph{Text classification} We use the lifelong text classification benchmark introduced by \citet{deMasson-episodic_memory} which consists of five datasets\footnote{\url{https://tinyurl.com/y89zdadp}} from \citet{zhang-text_cls}, trained on sequentially. The datasets are AGNews (news classification; $4$ classes), Yelp (sentiment analysis; $5$ classes), Amazon (sentiment analysis; $5$ classes), DBpedia (Wikipedia article classification; $14$ classes) and Yahoo (questions and answers categorization; $10$ classes). Following \citet{deMasson-episodic_memory}, we merge the classes of Yelp and Amazon and have a total of $33$ classes, and randomly sample $115,000$ training examples and $7,600$ test examples from each of the datasets since each of them have different sizes. The evaluation metric is the macro average of the accuracies over the five datasets.

\paragraph{Relation extraction} We use the lifelong relation extraction benchmark created by \citet{wang-embedding_alignment} based on the few-shot relation classification dataset FewRel \citep{han-fewrel}. It consists of $44,800$ training sentences and $11,200$ test sentences, and a total of $80$ relations along with their corresponding names available. Each sentence has a ground-truth relation as well as a set of $10$ negative candidate relations. The goal is to predict the correct relation among them. To construct tasks for continual learning, they first perform K-means clustering over the average GloVe embeddings \citep{pennington-glove} of the relation names to obtain $10$ disjoint clusters. Each task then comprises of data points having ground-truth relations from the corresponding cluster. In any given task, the candidate relations that were not seen in earlier tasks are removed. But, if all the candidate relations are unseen, the last two candidates are retained. The evaluation metric is the accuracy on a single test set containing relations from all the clusters. 

\subsection{Implementation}

For text classification, we largely maintain the experimental setup of \citet{deMasson-episodic_memory}. We consider four orders of the datasets (see Appendix \ref{sec:text_order}) and report the average results obtained from three independent runs. We also set $p_{write} = 1$. While they perform replay by drawing $100$ examples from memory for every $10,000$ examples from the stream, we draw $96$ examples from memory for every $9,600$ examples which is more convenient with batch size $b = 16$. Thus, we have $r = 0.01$ and $R_I = 9600$. We obtain the best hyperparameters by tuning on the first order of the datasets only. The learning rate for SEQ, A-GEM, REPLAY and MTL is $\expnumber{3}{-5}$. MTL is trained for $2$ epochs. For OML-ER, the inner-loop and outer-loop learning rates are $\expnumber{1}{-3}$ and $\expnumber{1}{-5}$ respectively whereas for ANML-ER, they are $\expnumber{3}{-3}$ and $\expnumber{1}{-5}$ respectively. The support set buffer size $m$ for both of them is $5$. We truncate the input sequence length to $300$ for ANML-ER and $448$ for the rest. The loss function is the cross-entropy loss across the $33$ classes. For the evaluation of meta-learning methods, we construct five episodes at meta-test time, one for each of the datasets, where their query sets consist of the test sets of these datasets. 

For relation extraction, we consider five orders of the tasks as in \citet{wang-embedding_alignment}. We report the average accuracy on the test set over the five orders, averaged over three independent runs. Sentence-relation pairs are concatenated with a [SEP] token between them to serve as the input. Since this is a smaller dataset, we set $R_I = 1600$ and $r = 0.01$. Additionally, $b = 4$, $m = 5$ and $p_{write} = 1$. Hyperparameter tuning is performed only on the first order. The learning rate is $\expnumber{3}{-5}$ for SEQ, A-GEM and REPLAY. MTL is trained with a learning rate of $\expnumber{1}{-5}$ for $3$ epochs. The inner-loop and outer-loop learning rates are $\expnumber{1}{-3}$ and $\expnumber{3}{-5}$ for OML-ER as well as ANML-ER. All models are trained using the binary cross-entropy loss, treating the true sentence-relation pairs as the positive class and the incorrect pairs as the negative class. The prediction is obtained as an argmax over the logit scores. Meta-learning methods are evaluated using a single meta-test episode with the test set as the query set.

\section{Experiments and results}

\paragraph{Text classification}

\begin{table*}[ht]
    \small
    \centering
    \begin{tabular}{llllll}
        \toprule
        \multicolumn{1}{c}{\multirow{2}{*}{\textbf{Method}}} & \multicolumn{5}{c}{\textbf{Accuracy}} \\
        & \makecell{Order 1} & \makecell{Order 2} & \makecell{Order 3} & \makecell{Order 4} & \makecell{Average} \\
        \midrule
        MbPA++ \citep{deMasson-episodic_memory} & 70.8 & 70.9 & 70.2 & 70.7 & 70.6 \\
        MbPA++ \citep{sun-lamol} & 74.1 & 74.9 & 73.1 & 74.9 & 74.2 \\
        LAMOL \citep{sun-lamol} & 76.7 & 77.2 & 76.1 & 76.1 & 76.5 \\
        Meta-MbPA \citep{wang-efficient} & 77.9 & 76.7 & 77.3 & 77.6 & 77.3 \\
        \cdashlinelr{1-6}
        SEQ & 16.7 $\pm$ 0.7 & 25.0 $\pm$ 0.5 & 19.5 $\pm$ 0.4 & 22.1 $\pm$ 0.5 & 20.8 $\pm$ 0.5 \\
        A-GEM & 16.6 $\pm$ 0.9 & 25.9 $\pm$ 1.1 & 21.6 $\pm$ 0.8 & 23.5 $\pm$ 1.0 & 21.9 $\pm$ 0.3 \\ 
        REPLAY & 69.5 $\pm$ 1.0 & 66.2 $\pm$ 2.0 & 65.2 $\pm$ 2.3 & 68.3 $\pm$ 2.2 & 67.3 $\pm$ 0.7 \\
        OML-ER & 75.4 $\pm$ 0.3 & \textbf{76.5 $\pm$ 0.2} & 75.4 $\pm$ 0.5 & 75.4 $\pm$ 0.8 & 75.7 $\pm$ 0.4 \\
        ANML-ER & \textbf{75.6 $\pm$ 0.4} & 75.8 $\pm$ 0.1 & \textbf{75.5 $\pm$ 0.3} & \textbf{75.7 $\pm$ 0.3} & \textbf{75.7 $\pm$ 0.1} \\
        \midrule
        MTL & --- & --- & --- & --- & 79.4 $\pm$ 0.2 \\
        \bottomrule
    \end{tabular}
    \caption[Test set accuracy on text classification]{Test set accuracy on text classification. The last column is the macro average across the four orders.}
    \label{tab:text_cls_results}
\end{table*}

We present the average accuracy across the baselines and our models with standard deviations across runs in Table \ref{tab:text_cls_results}. We perform significance testing with a two-tailed paired t-test at a significance level of $0.05$. Simply training on the datasets sequentially leads to extreme forgetting as reflected in the low accuracy of the SEQ model. With A-GEM, we get only a small, but significant gain ($p = 0.008$) compared to sequential training. By analyzing the frequency of constraint violations in Appendix \ref{sec:constraint_violations}, we find that A-GEM updates on BERT often behave similar to that in SEQ, which explains its poor performance. REPLAY, on the other hand, drastically improves performance, indicating that BERT benefits substantially even from a sparse experience replay. MbPA++ is the current state-of-the-art on this benchmark under the realistic setup of excluding task identifiers, using sparse replay and a single training epoch. \citet{sun-lamol} re-implement MbPA++ and obtain a higher score than the original implementation. We surmise that this is partly attributed to the fact that they perform replay after every $100$ steps along with dynamic batching and therefore likely resulting in a higher replay interval. Our approach, ANML-ER, achieves the highest accuracy, demonstrating that our meta-learning setup is more effective at mitigating catastrophic forgetting. OML-ER is almost as effective as ANML-ER, with the differences between the two being statistically insignificant ($p = 0.993)$. Although LAMOL has a higher score, it is not directly comparable to our methods since it uses task identifiers and multiple epochs of training, and has a higher generative replay rate of $20$\%, all of which make the task easier. Meta-MbPA is not directly comparable either since it performs local adaptation on nearest neighbors obtained from the memory during all its inner loop updates, thus having a higher replay rate effectively. Our meta-learning approach further narrows the gap with the MTL upper bound. 

\paragraph{Relation extraction}

\begin{table}[ht]
    \small
    \centering
    \begin{tabular}{ll}
        \toprule
        \makecell{\textbf{Method}} & \makecell{\textbf{Accuracy}} \\
        \midrule
        EA-EMR \citep{wang-embedding_alignment} & 56.6 \\
        MLLRE \citep{obamuyide-vlachos-2019-rel-extract} & 60.2 \\
        EMAR \citep{han-continual_rel} & 66.0 \\
        \cdashlinelr{1-2}
        SEQ & 48.1 $\pm$ 3.2 \\
        A-GEM & 45.5 $\pm$ 2.1 \\
        REPLAY & 65.4 $\pm$ 1.2 \\
        OML-ER & \textbf{69.5 $\pm$ 0.5} \\
        ANML-ER & 68.5 $\pm$  0.7 \\
        \midrule
        MTL & 85.7 $\pm$ 1.1 \\
        \bottomrule
    \end{tabular}
    \caption{Test set accuracy on relation extraction.}
    \label{tab:rel_results}
\end{table}

We report the average test set accuracy along with the standard deviation across the three runs in Table \ref{tab:rel_results}. We see that A-GEM performs similar to SEQ, with the differences being statistically insignificant ($p = 0.218$). Including sparse experience replay (REPLAY) again leads to a substantial increase in performance compared to SEQ. A low A-GEM performance compared to a simple replay method on this benchmark was also observed in \citet{wang-embedding_alignment}. OML-ER and ANML-ER significantly outperform all the baselines ($p = 0.006$ for OML-ER and $p = 0.026$ for ANML-ER when compared to REPLAY), and the former achieves the highest accuracy overall but, again, the differences between the two are not statistically significant ($p = 0.098$). Although not directly comparable, both of them outperform the previous state-of-the-art LSTM-based method EMAR \citep{han-continual_rel}, despite it using task identities as additional information and training for multiple epochs. There is, however, a wide gap between OML-ER and the MTL upper bound. We return to this in a later analysis. 

\section{Analysis}

\paragraph{Ablation study}

To investigate the relative strengths of the various components in our approach, we perform an ablation study and report the results in Table \ref{tab:ablation}. Meta-learning without replay leads to a large drop in performance, showing that experience replay, despite being sparse, is crucial. Interestingly however, meta-learning without replay still has considerably higher scores compared to SEQ, demonstrating that it is more resilient to catastrophic forgetting. Retrieving relevant examples from memory and fine-tuning on them during inference is a key aspect in MbPA++ since retrieving random examples instead produces only about $0.4\%$ improvement over REPLAY \citep{deMasson-episodic_memory}. Our approach works with random examples and yet achieves substantially higher accuracies. For OML-ER, omitting fine-tuning altogether at the meta-testing stage produces a small, yet significant drop ($p = 0.019$) for text classification, but an insignificant one ($p = 0.602$) for relation extraction. Similarly, for ANML-ER, no meta-test fine-tuning results in a small, significant drop ($p = 0.006$) for text classification and an insignificant change ($p = 0.265$) for relation extraction. Unlike MbPA++, our methods, overall, work well even without additional adaptation steps during inference. Without neuromodulation, ANML-ER is equivalent to standard MAML enhanced with experience replay, which we could call MAML-ER. The performance difference between ANML-ER and MAML-ER is not statistically ($p = 0.120$ for text classification and $p = 0.087$ for relation extraction), which suggests that the neuromodulator in ANML-ER is not useful for our language tasks. Even though OML-ER, ANML-ER and MAML-ER are equally successful in terms of performance, OML-ER is computationally more efficient as only its PLN (a single linear layer) is fine-tuned in the inner-loop.

\begin{table}[ht]
    \small
    \centering
    \begin{tabular}{lcc}
        \toprule
        \multicolumn{1}{c}{\multirow{3}{*}{\textbf{Method}}} & \multicolumn{2}{c}{\textbf{Accuracy}} \\
        & \makecell{Text \\ classification} & \makecell{Relation \\ extraction} \\
        \midrule
        OML-ER & 75.7 $\pm$ 0.4 & 69.5 $\pm$ 0.5 \\
        \quad $-$ Replay & 24.6 $\pm$ 0.6 & 55.9 $\pm$ 0.9 \\
        \quad $-$ Meta-test fine-tuning & 75.6 $\pm$ 0.4 & 69.3 $\pm$ 0.7 \\
        \midrule
        ANML-ER & 75.7 $\pm$ 0.1 & 68.5 $\pm$  0.7 \\
        \quad $-$ Replay & 51.7 $\pm$ 1.8 & 57.0 $\pm$ 0.9 \\
        \quad $-$ Meta-test fine-tuning & 74.9 $\pm$ 0.3 & 67.7 $\pm$ 0.9 \\
        \quad $-$ Neuromodulation & 75.8 $\pm$ 0.2 & 68.0 $\pm$ 0.4 \\
        \bottomrule
    \end{tabular}
    \caption{Ablation study on model components.}
    \label{tab:ablation}
\end{table}

\paragraph{Effect of replay rate}

\begin{table}[ht]
    \small
    \centering
    \begin{tabular}{clcc}
        \toprule
        \multicolumn{1}{c}{\multirow{3}{*}{\textbf{Replay rate}}} & \multicolumn{1}{c}{\multirow{3}{*}{\textbf{Method}}} & \multicolumn{2}{c}{\textbf{Accuracy}} \\
        & & \makecell{Text \\ classification} & \makecell{Relation \\ extraction} \\
        \midrule
        \multirow{2}{*}{1 \%} & REPLAY & 67.3 $\pm$ 0.7 & 65.4 $\pm$ 1.2 \\
         & OML-ER & 75.7 $\pm$ 0.4 & 69.5 $\pm$ 0.5 \\
         \midrule
        \multirow{2}{*}{2 \%} & REPLAY & 67.2 $\pm$ 2.0 & 67.1 $\pm$ 0.8 \\
         & OML-ER & 75.6 $\pm$ 0.1 & 71.6 $\pm$ 1.1 \\
          \midrule
        \multirow{2}{*}{4 \%} & REPLAY & 70.3 $\pm$ 1.3 & 69.2 $\pm$ 2.2 \\
         & OML-ER & 76.0 $\pm$ 0.6 & 75.5 $\pm$ 0.4 \\
        \midrule
        --- & MTL & 79.4 $\pm$ 0.2 & 85.7 $\pm$ 1.1 \\
        \bottomrule
    \end{tabular}
    \caption{Test metrics on text classification and relation extraction for varying replay rates.}
    \label{tab:replay_rate_exp}
\end{table}

We noted previously that there exists a gap in performance between our best model and MTL. To analyze if increasing the replay rate can help narrow the gap, we train both REPLAY and OML-ER with a $2$\% and $4$\% replay rate\footnote{The maximum replay rate for our meta-learning methods is $1/m = 20$\% i.e., replay every episode with $m=5$}, keeping $R_I$ the same as before (Table \ref{tab:replay_rate_exp}). On text classification, OML-ER has similar performance ($p = 0.936$) with $2$\% replay rate and a small, significant improvement ($p = 0.001$) with $4$\% replay rate. The same trend is observed with REPLAY as the replay rate increases ($p = 0.861$ and $p = 0.045$) . In contrast, OML-ER and REPLAY improve by a significantly greater extent on relation extraction ($p = \expnumber{3}{-4}$ and $p = 0.048$ respectively). We surmise this is because text classification has equally sized tasks whereas the tasks in relation extraction are imbalanced (see Appendix \ref{sec:rel_dist}). Since we employ uniform sampling for memory read/write, this imbalance is reflected in the memory, causing larger tasks to be replayed more often and underrepresented tasks to be forgotten more quickly. A higher replay rate therefore increases the chances of sufficiently revisiting all previous tasks, leading to better scores. Additionally, on both benchmarks, OML-ER outperforms REPLAY even with higher replay rates. There is still a wide gap between OML-ER with $4$\% replay and MTL, indicating there is scope for improvement. 

\paragraph{Effect of memory size}

\begin{table}[ht]
    \small
    \centering
    \begin{tabular}{ccc}
        \toprule
        \multirow{2}{*}{\makecell{\textbf{Memory} \\ \textbf{capacity}}} & \multicolumn{2}{c}{\textbf{Accuracy}} \\
        & \makecell{Text classification} & \makecell{Relation extraction} \\
        \midrule
        100 \% & 75.7 $\pm$ 0.4 & 69.5 $\pm$ 0.5 \\
        5 \% & 75.2 $\pm$ 0.4 & 69.2 $\pm$ 0.9 \\
        1 \% & 75.6 $\pm$ 0.3 & 66.8 $\pm$ 0.3 \\
        \bottomrule
    \end{tabular}
    \caption{Variation of performance of OML-ER with the size of the memory.}
    \label{tab:memory_exp}
\end{table}

In our experiments so far, we store all the examples in memory; however, this does not scale well when the number of tasks is very large. In order to investigate the effect of memory size on performance, we present the accuracy of OML-ER with $5$\% and $1$\% memory capacity in Table \ref{tab:memory_exp}. We achieve this by setting $p_{write}$ to $0.05$ and $0.01$ respectively. There are insignificant changes ($p = 0.074$ and $p = 0.952$ respectively) in average accuracy even with reduced memory for text classification. MbPA++, on the other hand, was shown to have a drop of $3$\% accuracy with $10$\% memory capacity \citep{deMasson-episodic_memory}, which demonstrates that our method is more memory-efficient. Performance on relation extraction suffers a small but significant drop ($p = 0.040$) with $1$\% memory. The difference is insignificant ($p = 0.741$) with $5$\% memory and, overall, can still be considered memory-efficient. 

\paragraph{Episodic updates} In addition to the automatic gradient alignment that comes with meta-learning, we believe that its episodic nature is another reason for its strength in lifelong learning. In text classification for example, SEQ has a replay every $600$ optimizer steps whereas meta-learning, by way of its formulation, has a replay every $101$ meta-optimizer steps (using Equation \ref{eqn:replay_frequency} with our hyperparameters)\footnote{However, we note that optimizer steps and meta-optimizer steps are not the same nor directly comparable as such.}. Fewer updates between replays likely aids in knowledge retention. To probe deeper, we trained our REPLAY model such that replay occurs every $100$ optimizer steps by setting $R_I = 1600$, with everything else being the same. This achieves an accuracy of $74.4 \pm 0.2$. Although this is now closer to our meta-learning methods, it is still significantly lower ($p = 0.001$ for OML-ER and $p = \expnumber{8}{-5}$ for ANML-ER). Therefore, episodic updates in meta-learning are an important part of the model, contributing positively to performance. For ``regular'' training to match the same level of performance, experience replay would need to be performed more often. 

\section{Discussion}
Continual learning methods so far have relied on manual heuristics and/or have computational bottlenecks. MbPA++ is inexpensive during training due to sparse replay, but its inference is expensive since it requires retrieval of $K$ nearest neighbors for every test example and multiple gradient steps on them. A-GEM, on the other hand, is slower to train due to its projection steps. OML-ER achieves the best of both worlds -- its training is fast because its inner-loop, which makes up a large portion of the training, involves only updating the small PLN, and its inference is fast since it relies only on a small number of updates on randomly drawn examples from memory. Furthermore, it also retains its performance when the memory capacity is scaled down. 

Our method uses a simple, random write mechanism. Other strategies such as those based on surprise \citep{ramalho-surprise} and forgetting \citep{toneva-forgetting} could further refine performance. Furthermore, the problem of task size imbalance could be mitigated with class-balancing reservoir sampling \citep{chrysakis_continual}.

In our experiments on text classification, we assume that all the classes are known beforehand. Lifelong learning when the classes are unknown \textit{a priori} and available only during each of the individual tasks is more challenging and would be an interesting extension.

Recently, \citet{knoblauch-nphard} showed theoretically that optimal continual learning is an NP-hard problem and requires perfect memorization of the past. An implication of this finding is that replay-based methods are more effective than regularization-based methods. Therefore, experience replay would perhaps remain a key component in designing future, more advanced methods. 

Another promising direction for future work would be to integrate differential Hebbian plasticity \citep{miconi-plasticity,miconi-backpropamine} into a meta-learning framework for continual learning. Designing an appropriate neuromodulator for transformer-based language models or encouraging sparsity in them also requires additional work. 
 
\section{Conclusion}
We showed that pre-trained transformer-based language models, meta-learning and sparse experience replay produce a synergy that improves lifelong learning on language tasks in a realistic setup. This is an important step in moving away from manually-designed solutions into simpler, more generalizable methods to ultimately achieve human-like learning. Meta-learning could further be exploited for the combined setting of few-shot and lifelong learning. It might also be promising in learning distinct NLP tasks in a curriculum learning fashion.  

\bibliography{main}
\bibliographystyle{acl_natbib}

\appendix

\section{Appendix}

\subsection{Dataset order for text classification} \label{sec:text_order}
For text classification, the four different orderings of the datasets are:
\begin{enumerate}[itemsep=0mm]
\small
\item Yelp $\rightarrow$ AGNews $\rightarrow$ DBpedia $\rightarrow$ Amazon $\rightarrow$ Yahoo
\item DBpedia $\rightarrow$ Yahoo $\rightarrow$ AGNews $\rightarrow$ Amazon $\rightarrow$ Yelp
\item Yelp $\rightarrow$ Yahoo $\rightarrow$ Amazon $\rightarrow$ DBpedia $\rightarrow$ AGNews
\item AGNews $\rightarrow$ Yelp $\rightarrow$ Amazon $\rightarrow$ Yahoo $\rightarrow$ DBpedia
\end{enumerate}

\subsection{Task distribution for relation extraction} \label{sec:rel_dist}
\begin{figure}[ht]
    \centering
    \begin{subfigure}[b]{0.46\columnwidth}
         \centering
         \includegraphics[width=\linewidth]{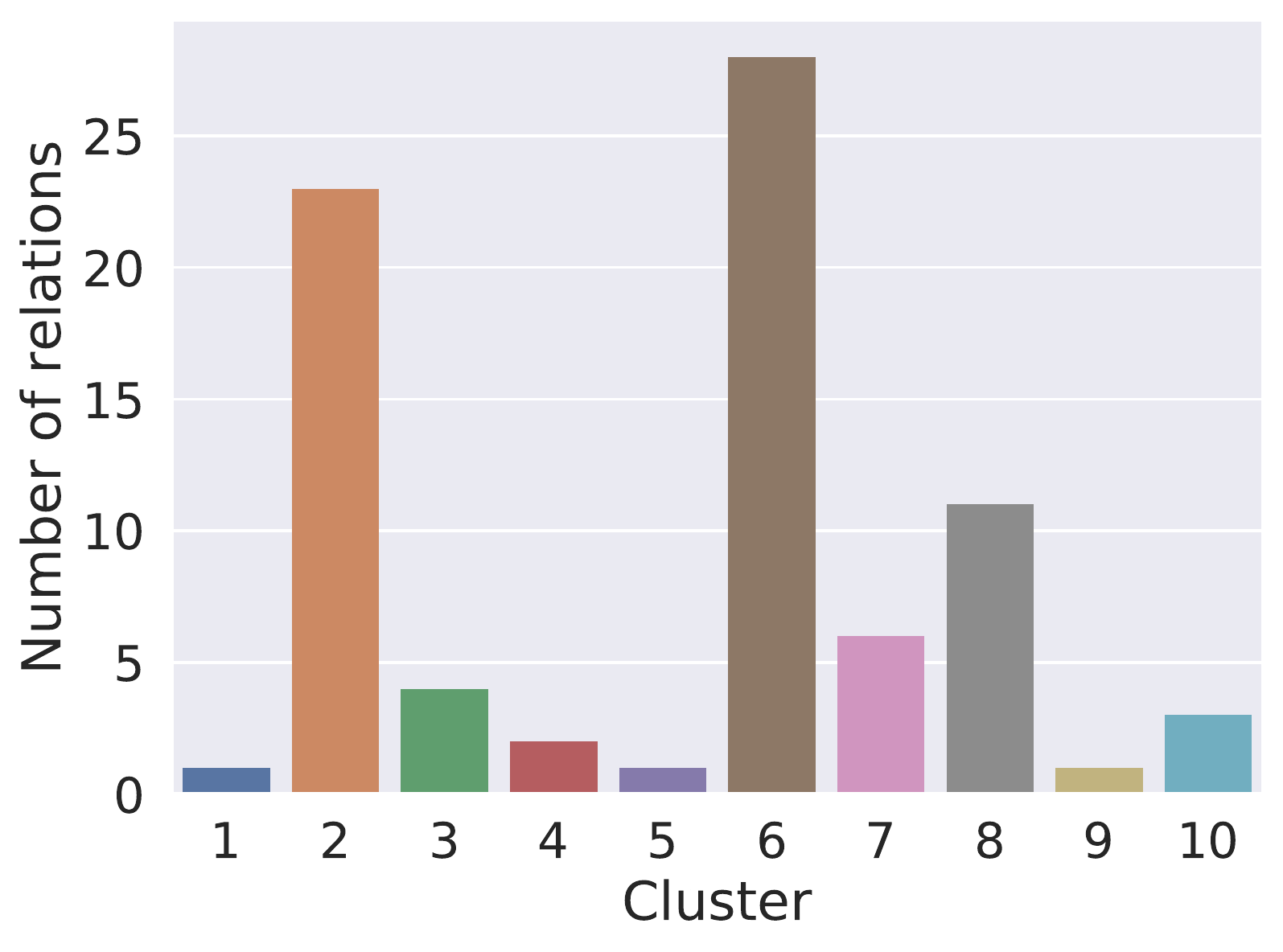}
         \caption{Number of relations per cluster.}
         \label{fig:cluser_nrels}
     \end{subfigure}
     \hfill
     \begin{subfigure}[b]{0.49\columnwidth}
         \centering
         \includegraphics[width=\linewidth]{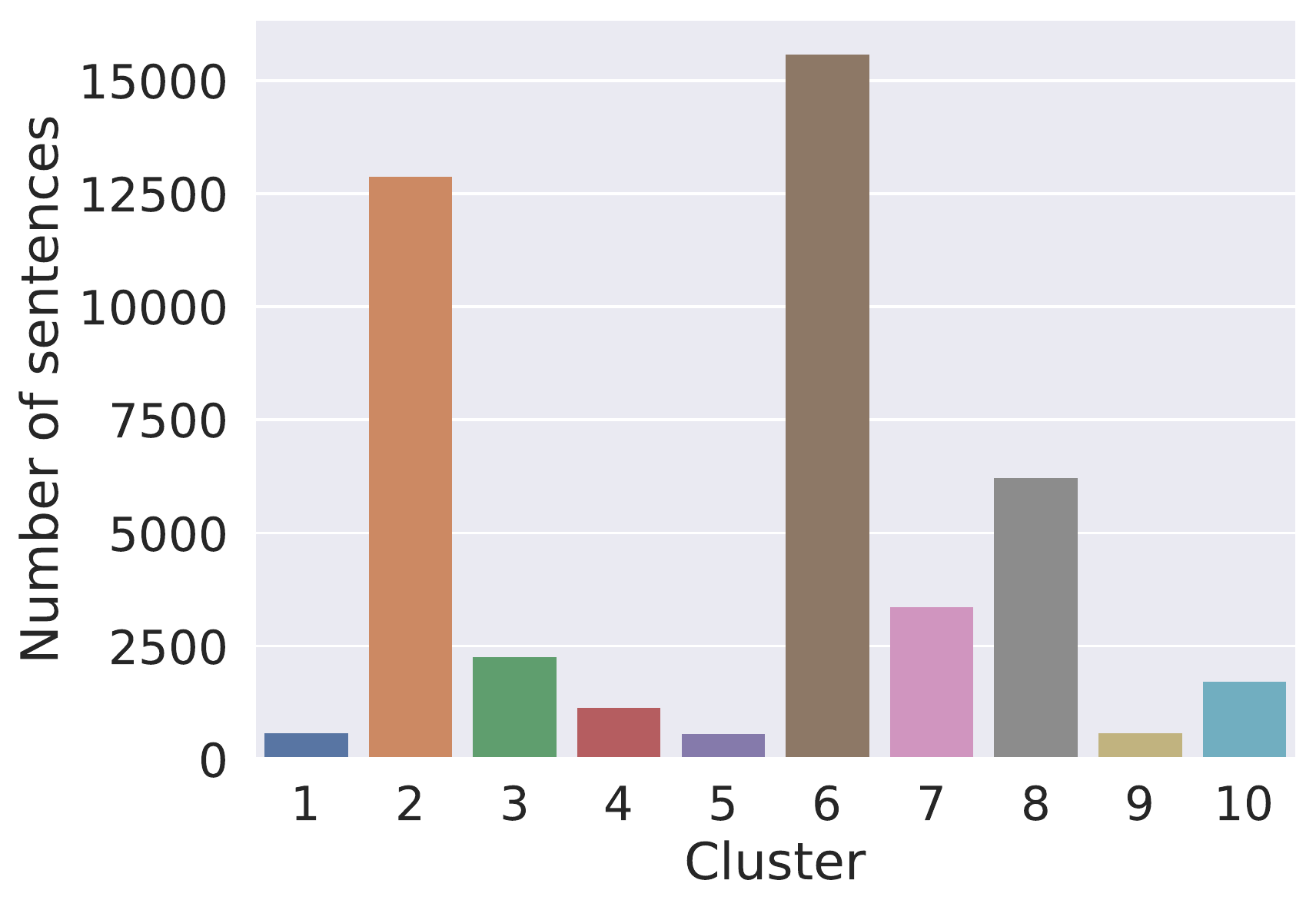}
         \caption{Number of sentences per cluster.}
         \label{fig:cluster_sizes}
     \end{subfigure}
    \caption{Task distribution for relation extraction.}
    \label{fig:rel_dist}
\end{figure}

In relation extraction, the size of each cluster is not balanced. Hence, each of the tasks vary in their size. In Figure \ref{fig:rel_dist} we plot the number of relations and the number of sentences in each cluster. Overall, there is a great imbalance with respect to the task size, with cluster 2 and 6 having a disproportionately larger size compared to the other clusters.  

\subsection{Expression for replay frequency} \label{sec:RF}
In REPLAY and A-GEM, since gradient updates occur after seeing a batch of size $b$ from the stream, the replay frequency $R_F$, i.e., the number of steps between the replay interval $R_I$, is simply given by
\begin{equation}
    R_F = \left \lceil \frac{R_I}{b} \right \rceil 
\end{equation}
In meta-learning, learning occurs in episodes where the support set has $m$ batches of size $b$ each and a single batch as query set of the same size $b$. After encountering $R_I$ examples, we would like the replay to be realized as a query set. If $R_F$ is the episode at which replay occurs, 
\begin{gather}
    b \left[ (R_F - 1) (m + 1) + m \right] = R_I \nonumber \\
    R_F - 1 = \frac{R_I / b - m}{m + 1} \nonumber \\
    R_F = \left \lceil \frac{R_I / b + 1}{m + 1} \right \rceil
\end{gather}
where we round it up to the nearest integer so that replay is not performed before $R_I$ examples. 

\subsection{Derivation of gradients} \label{sec:grad_deriv}
Consider a first-order MAML setup that performs one step of SGD on each of the $m$ batches in the support set during the inner-loop of an episode. Starting with parameters $\bm{\theta}_0 = \bm{\theta}$, it results in a sequence of parameters $\bm{\theta}_1$, ..., $\bm{\theta}_m$ using the losses $\mathcal{L}^1$, ..., $\mathcal{L}^m$. The query set could be considered as the $(m+1)$-th batch that produces the meta-gradient for $\bm{\theta}$ using $\mathcal{L}^{(m+1)} = \mathcal{L}^q$. We introduce the following two notations to denote the gradient and the Hessian with respect to the initial parameters $\bm{\theta}$:
\begin{align}
    \bar{g}_i &= \frac{\partial \mathcal{L}^i(\bm{\theta}_{i-1})}{\partial \bm{\theta}} \label{eqn:grad_notation}\\
    \bar{H}_i &= \frac{\partial^2 \mathcal{L}^i(\bm{\theta}_{i-1})}{\partial \bm{\theta}^2} \label{eqn:hessian_notation}
\end{align}

Using Taylor series approximation, \citet{Nichol} show that the meta-gradient can be written as:
\begin{align}
    g_{\text{FOMAML}} &= \frac{\partial \mathcal{L}^q(\bm{\theta}_m)}{\partial \bm{\theta}_m} \nonumber \\
    &= \bar{g}_{m+1} - \alpha \bar{H}_{m+1} \sum_{j=1}^m \bar{g}_j + O(\alpha^2) \nonumber
\end{align}
Taking expectation under mini-batch sampling, 
{\small 
\begin{align}
    \mathbb{E}[g_{\text{FOMAML}}] &= \mathbb{E} \left[ \bar{g}_{m+1} \right] - \alpha  \sum_{j=1}^m \mathbb{E} \left[ \bar{H}_{m+1} \bar{g}_j \right] +  O(\alpha^2) \nonumber \\ 
    &= \mathbb{E} \left[ \bar{g}_{m+1} \right] - \alpha  \sum_{j=1}^m \mathbb{E} \left[ \bar{H}_{j} \bar{g}_{m+1} \right] +  O(\alpha^2) \nonumber \\
    & \quad \text{(interchanging $j$ and $m+1$)} \nonumber \\
    &= \mathbb{E} \left[ \bar{g}_{m+1} \right] - \frac{\alpha}{2}  \sum_{j=1}^m \mathbb{E} \left[ \bar{H}_{m+1} \bar{g}_j + \bar{H}_{j} \bar{g}_{m+1} \right] \nonumber \\
    & \quad + O(\alpha^2) \; \text{(averaging the last two equations)} \nonumber \\
    &= \mathbb{E} \left[ \bar{g}_{m+1} \right] - \frac{\alpha}{2} \frac{\partial}{\partial \bm{\theta}} \sum_{j=1}^m \mathbb{E} \left[ \bar{g}_j \cdot \bar{g}_{m+1} \right] \nonumber \\
    & \quad +  O(\alpha^2) \nonumber
\end{align}}
\normalsize
Re-writing based on Equation \ref{eqn:grad_notation} and \ref{eqn:hessian_notation} gives:

{\small
\begin{dmath}
    \mathbb{E}[g_{\text{FOMAML}}] = \mathbb{E} \left[ \frac{\partial \mathcal{L}^q(\bm{\theta}_{m})}{\partial \bm{\theta}} - \frac{\alpha}{2} \frac{\partial}{\partial \bm{\theta}} \left( \sum_{j=1}^m \frac{\partial \mathcal{L}^j(\bm{\theta}_{j-1})}{\partial \bm{\theta}} \cdot \frac{\partial \mathcal{L}^q(\bm{\theta}_{m})}{\partial \bm{\theta}} \right) \right] +  O(\alpha^2) 
\end{dmath}}
\normalsize

\subsection{Implementation details}

For text classification, we take $5,000$ examples from each of the datasets as the validation set and $115,000$ examples from each of the datasets as the training set. The number of training examples matches that of \citet{deMasson-episodic_memory}. On the other hand, for relation extraction, we take a subset of $4,800$ examples from the training set as the validation set for hyperparameter tuning. With the best hyperparameters, we re-train on all $44,800$ examples to match the number of examples used in \citet{wang-embedding_alignment}.

The only hyperparameters we tune are the learning rate (for SEQ, A-GEM and REPLAY), the inner and meta learning rates, and the support set buffer size $m$ (for OML-ER and ANML-ER). The other hyperparameters are fixed to appropriate values. We performed tuning over the following values:
\begin{itemize}\itemsep0em
    \item Learning rate: $\expnumber{5}{-4}$, $\expnumber{1}{-5}$, $\expnumber{3}{-5}$, $\expnumber{5}{-5}$
    \item Inner learning rate: $\expnumber{5}{-2}$, $\expnumber{1}{-3}$, $\expnumber{3}{-3}$, $\expnumber{5}{-3}$
    \item Meta learning rate: $\expnumber{5}{-4}$, $\expnumber{1}{-5}$, $\expnumber{3}{-5}$, $\expnumber{5}{-5}$
    \item $m$: $3$, $5$, $7$, $9$
\end{itemize}

In Table \ref{tab:hyperparams}, we summarize all the hyperparameters for text classification and relation extraction. We use the random seeds $42$ -- $44$ for the three independent runs. All models were trained on a system with a single Nvidia Titan RTX GPU and $45$ GB memory.

\begin{table*}[ht]
    \small
    \centering
    \begin{tabular}{lcccccc} 
    \toprule
    \textbf{Model} & \textbf{Learning rate} & \makecell{\textbf{Inner loop} \\ \textbf{learning rate}} & \makecell{\textbf{Meta} \\ \textbf{learning rate}} & \makecell{\textbf{Support set} \\ \textbf{buffer size}} & \makecell{\textbf{Batch size}} & \makecell{\textbf{Maximum} \\ \textbf{sequence length}} \\
    \midrule 
    SEQ & $\expnumber{3}{-5}$ & --- & --- & --- & 16 & 448\\
    A-GEM & $\expnumber{3}{-5}$ & --- & --- & --- & 16 & 448 \\
    REPLAY & $\expnumber{3}{-5}$ & --- & --- & --- & 16 & 448  \\
    MTL (2 epochs) & $\expnumber{3}{-5}$ & --- & --- & --- & 16 & 448  \\
    OML-ER & --- & $\expnumber{1}{-3}$ & $\expnumber{1}{-5}$ & 5 & 16 & 448 \\
    ANML-ER & --- & $\expnumber{3}{-3}$ & $\expnumber{1}{-5}$ & 5 & 16 & 300 \\
    \midrule
    SEQ & $\expnumber{3}{-5}$ & --- & --- & --- & 4 & ---\\
    A-GEM & $\expnumber{3}{-5}$ & --- & --- & --- & 4 & --- \\
    REPLAY & $\expnumber{3}{-5}$ & --- & --- & --- & 4 & ---  \\
    MTL (3 epochs) & $\expnumber{1}{-5}$ & --- & --- & --- & 4 & ---  \\
    OML-ER & --- & $\expnumber{1}{-3}$ & $\expnumber{3}{-5}$ & 5 & 4 & --- \\
    ANML-ER & --- & $\expnumber{1}{-3}$ & $\expnumber{3}{-5}$ & 5 & 4 & --- \\
    \bottomrule         
    \end{tabular}
    \caption{Hyperparameters for text classification (top) and relation extraction (bottom).}
    \label{tab:hyperparams}
\end{table*}

\subsection{Frequency of constraint violations} \label{sec:constraint_violations}

\begin{figure}
    \centering
    \includegraphics[width=\columnwidth]{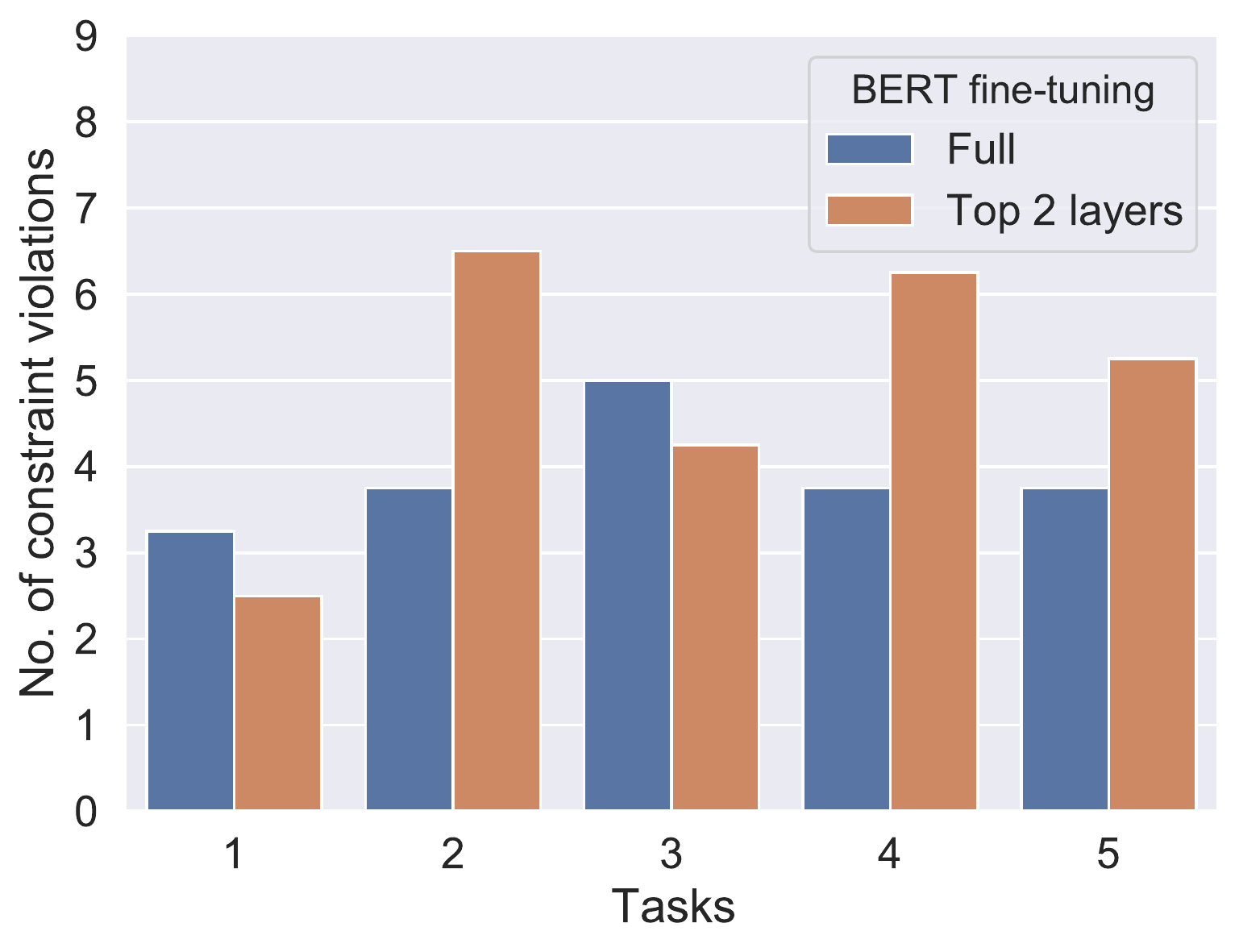}
    \caption{Average number of constraint violations per task in text classification.}
    \label{fig:violations}
\end{figure}

A-GEM solves a constrained optimization problem such that the dot product between the gradients from the current batch and a randomly drawn batch from the memory is greater than or equal to zero. We check constraint satisfaction by treating the model parameters as a single vector. To analyze the poor performance of A-GEM on our setup, we plot the average number of constraint violations across the four orders that occur per task in text classification in Figure \ref{fig:violations}. Note that the total number of optimizer steps per task is 7187 and replay occurs about 11 times for each. When fine-tuning the whole of BERT, we have relatively few violations, meaning that no gradient correction is done most of the time. This perhaps relates to the finding by \citet{merchant-bert_what_happens} that fine-tuning BERT primarily affects the top layers and does not lead to catastrophic forgetting of linguistic phenomena in the deeper layers. We see that the number of violations increase when we only fine-tune the top 2 layers of BERT. Yet, it was insufficient to reach the performance of a simple replay method.

\subsection{ANML visualization} 

\begin{figure*}
    \centering
    \includegraphics[width=\textwidth]{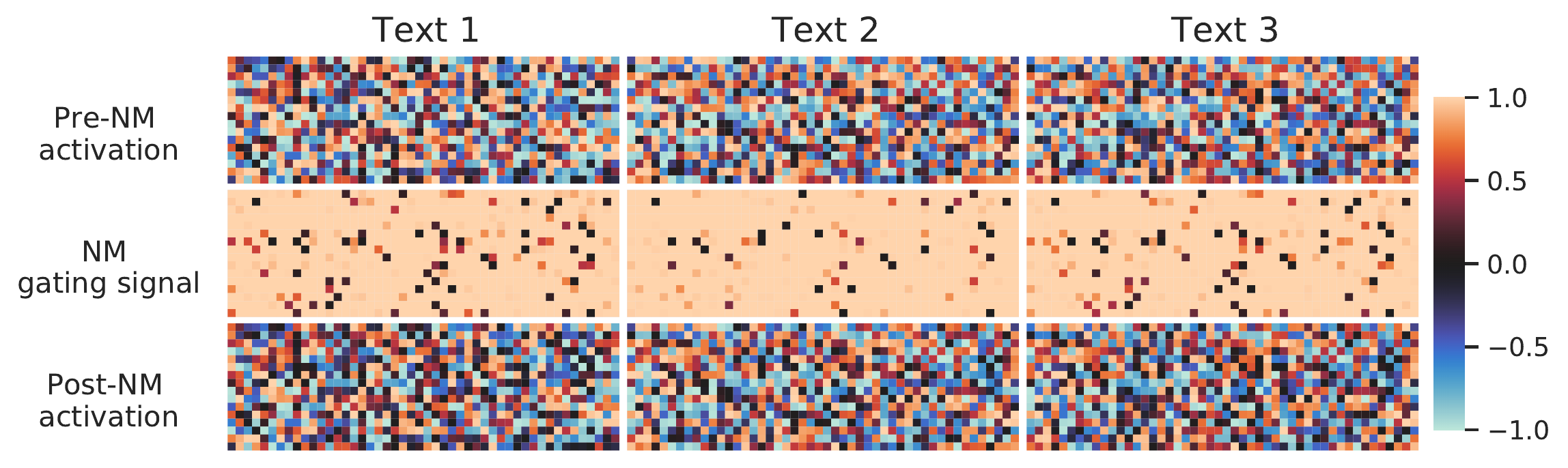}
    \caption{Visualization of the neuromodulatory signal (middle row) and the representation from BERT before (top row) and after (bottom row) neuromodulation for three randomly chosen texts from the AGNews dataset. We obtain the plots by reshaping the 768-dimensional representation into 48 $\times$ 16.}
    \label{fig:anml_visual}
\end{figure*}

The original OML and ANML models were shown to produce sparse representations with CNN encoders for images \citep{javed_oml,beaulieu-anml}. Sparse representations alleviate forgetting since only a few neurons are active for a given input. We visualize the representations from BERT before and after neuromodulation, along with the neuromodulatory signal, in our ANML-ER model in Figure \ref{fig:anml_visual}. Clearly, none of the representations are sparse. Moreover, most of the neuromodulatory signal is composed of ones, further confirming our hypothesis that the neuromodulator does not play a significant role here. The lack of sparsity was also observed in OML-ER. Perhaps, a more sophisticated neuromodulatory mechanism is required to induce sparsity in pre-trained transformer-based language models.

\end{document}